\makeatletter
\def\thanks#1{\protected@xdef\@thanks{\@thanks
        \protect\footnotetext{#1}}}
\makeatother
\documentclass[runningheads]{llncs}
\usepackage[T1]{fontenc}
\usepackage{graphicx}
\usepackage{amsmath}
\usepackage{multirow}  
\usepackage{multicol}  
\usepackage{amsfonts}
\usepackage{amssymb}
\usepackage{color}
\usepackage{xcolor}
\usepackage{hyperref}
\usepackage{marvosym}

\newcommand{\shuang}[1]{\textcolor{teal}{#1}}
\begin{document}
\title{Tackling Data Heterogeneity in Federated Learning via Loss Decomposition}
%
%
\author
{
Shuang Zeng\inst{1*} \and
Pengxin Guo \inst{1*}\and
Shuai Wang \inst{2} \and
Jianbo Wang \inst{3} \and
Yuyin Zhou \inst{4} \and
Liangqiong Qu \inst{1}\textsuperscript{(\Letter)}
\thanks{* These authors contributed equally to this work.}
\thanks{\textsuperscript{\Letter} Corresponding author.}
}
\authorrunning{S. Zeng et al.}
%
\institute
{
Department of Statistics and Actuarial Science, The University of Hong Kong, Hong Kong SAR, China \\
\email{liangqqu@hku.hk} \and
School of Cyberspace, Hangzhou Dianzi University, Hangzhou, China \and
New H3C Technologies Co., Ltd. \and
Computer Science and Engineering, University of California, Santa Cruz, USA 
}
\maketitle              
\begin{abstract}
Federated Learning (FL) is a rising approach towards collaborative and privacy-preserving machine learning where large-scale medical datasets remain localized to each client. However, the issue of data heterogeneity among clients often compels local models to diverge, leading to suboptimal global models. To mitigate the impact of data heterogeneity on FL performance, we start with analyzing how FL training influence FL performance by decomposing the global loss into three terms: local loss, distribution shift loss and aggregation loss. 
Remarkably, our loss decomposition reveals that existing local training-based FL methods attempt to reduce the distribution shift loss, while the global aggregation-based FL methods propose better aggregation strategies to reduce the aggregation loss. Nevertheless, a comprehensive joint effort to minimize all three terms is currently limited in the literature, leading to subpar performance when dealing with data heterogeneity challenges. 
To fill this gap, we propose a novel FL method based on global loss decomposition, called FedLD, to jointly reduce these three loss terms. Our FedLD involves a margin control regularization in local training to reduce the distribution shift loss, and a principal gradient-based server aggregation strategy to reduce the aggregation loss.
Notably, under different levels of data heterogeneity, our strategies achieve better and more robust performance on retinal and chest X-ray classification compared to other FL algorithms. Our code is available at \href{https://github.com/Zeng-Shuang/FedLD}{https://github.com/Zeng-Shuang/FedLD}.

\end{abstract}
\section{Introduction}
Federated Learning (FL), where computations are performed locally at each client without sharing data, presents a promising approach to accessing large, representative data for training robust deep learning models with enhanced generalizability~\cite{mcmahan2017communication}. In recent years, FL has witnessed some pivotal success on various medical applications, such as medical image segmentation  \cite{ziller2021medical}, medical image classification~\cite{dayan2021federated}, cancer boundary detection \cite{pati2022federated}, among others \cite{kaissis2020secure,dayan2021federated,kalra2023decentralized}. Despite its widespread, FL suffers from data heterogeneity \cite{yang2021characterizing,zhang2024flhetbench}, as data can be non-independent and identically distributed (non-IID) across clients, which is particularly prevalent in medical scenarios \cite{li2020multi,yan2023label,qu2021experimental,zhang2022splitavg,qu2022rethinking}.

Two main approaches have been proposed to tackle data heterogeneity in FL: (\textbf{i}) regularizing \textit{local training} to mitigate the deviation between local and global objectives and (\textbf{ii}) designing more efficient \textit{global aggregation} strategies. For local training, methods such as FedProx~\cite{li2020federated}, SCAFFOLD~\cite{karimireddy2020scaffold}, FedCM~\cite{xu2021fedcm}, and FEDIIR~\cite{guo2023out} utilize the difference between local and global models as a regularization term to mitigate the deviation between local and global objectives. However, these methods may suffer from additional communication costs~\cite{karimireddy2020scaffold} or the inability of gradient differences to accurately capture model bias~\cite{hamilton2017representation,xu2021fedcm,guo2023out}. 
For global aggregation, several works have been proposed to develop efficient global aggregation strategies, such as FedMA~\cite{wang2020federated}, pFedLA \cite{ma2022layer}, RobFedAvg \cite{uddin2021robust}, and GAMF \cite{liu2022deep}. However, these strategies may incur extra computing resources~\cite{liu2022deep,uddin2021robust}  or overlook the impact of local training on overall performance~\cite{charles2021large,zhang2023tackling}.
\par 
\begin{figure}[t]  
    \centering  
    \includegraphics[width=0.9\textwidth]{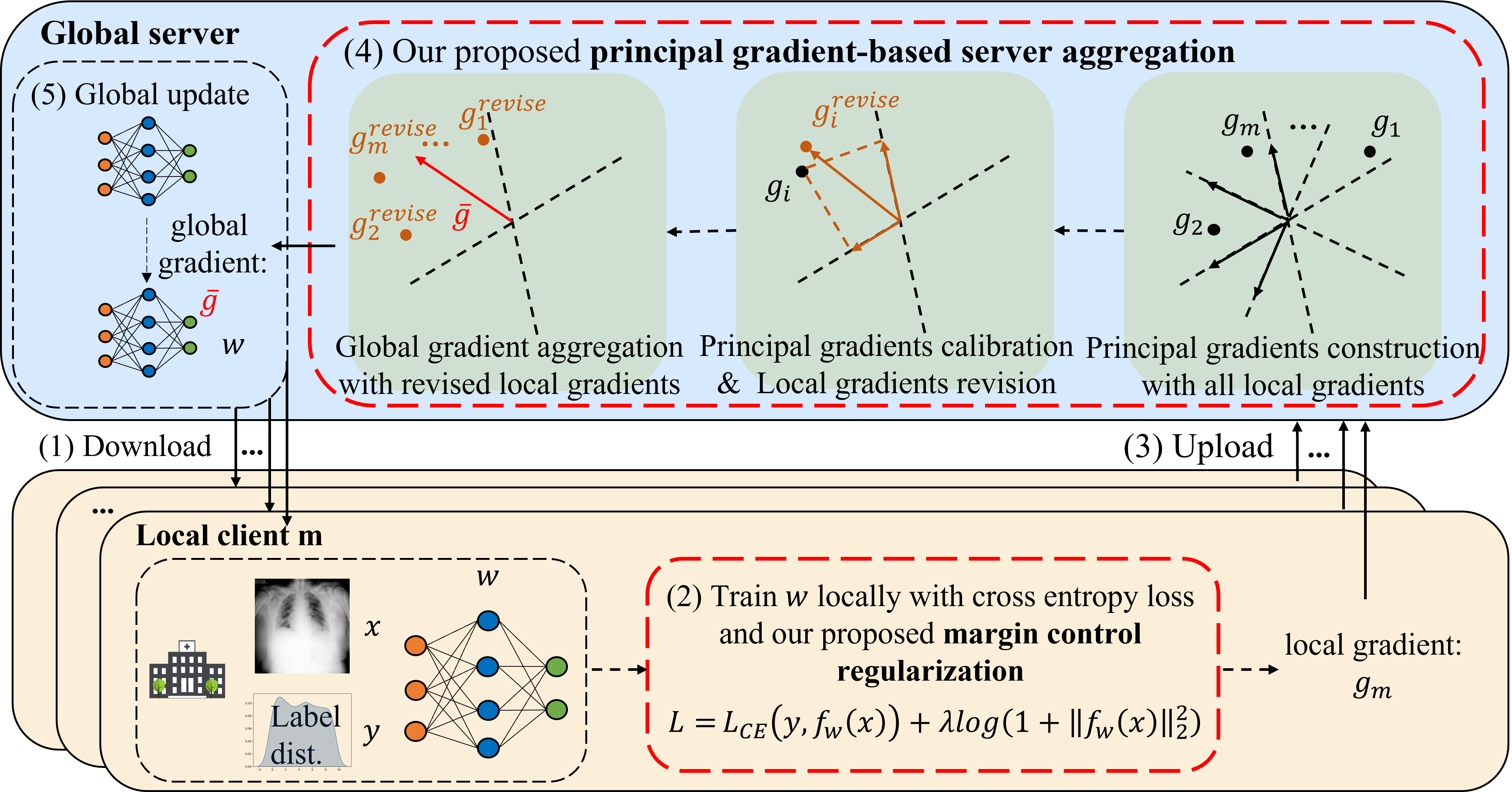}  
    \caption{Overview of the proposed FedLD. (1) Once each local client downloads the global model parameter $\boldsymbol{w}$, (2) it starts training locally with the cross-entropy loss and our proposed margin control regularization. (3) After that, each client uploads its local gradient to the global server. (4) Then, the global server aggregates these local gradients with our proposed principal gradient-based server aggregation, which includes three steps: First, use all local gradients to construct principal gradients; Second, calibrate principal gradients and use them to revise local gradients; Third, aggregate revised local gradients to generate the global gradient. (5) Finally, the server updates the global model parameter with the global gradient and sends it to local clients for the next round.} 
    \label{fig:framework}  
\vskip -0.2in
\end{figure} 

Most of the works address the issue of data heterogeneity by applying strategies either at the local or server side. In this paper, we ask: \textit{How can the joint effect of local training and server aggregation be leveraged to improve FL performance in the presence of data heterogeneity?} To answer this, we decompose the global loss into three terms: local loss, distribution shift loss, and aggregation loss, which reveals that existing methods often focus on addressing only one or two of these terms, rather than all three. For example, in standard FL training, such as FedAvg~\cite{mcmahan2017communication}, only the local loss is minimized, while the other two terms are ignored. Remarkably, existing local training-based methods attempt to further reduce the distribution shift loss through additional local training regularization \cite{li2020federated,karimireddy2020scaffold,xu2021fedcm,guo2023out}, and global aggregation-based methods propose better aggregation strategies to minimize the aggregation loss \cite{wang2020federated,ma2022layer,uddin2021robust,liu2022deep}.

However, a comprehensive joint effort to minimize all three terms - local loss, distribution shift loss, and aggregation loss - is currently limited in the literature. To fill this gap, we propose a novel FL method based on global loss decomposition, called FedLD, to jointly reduce these three loss terms, as shown in Fig.~\ref{fig:framework}. Our FedLD involves a margin control regularization in local training to reduce the distribution shift loss, and a principal gradient-based server aggregation strategy to minimize the aggregation loss. Specifically, our margin control regularization  encourages local models to learn stable features instead of shortcut features by adding the $l_2$-norm of the output logits to the standard cross entropy loss, thereby reducing the distribution shift loss. On the other hand, the local models trained on heterogeneous data distribution in FL may exhibit different or even conflicting judgments, leading to potential conflicts in clients' gradients. Naively aggregating these conflicting gradients in FL may lead to increased aggregation loss, thus resulting in poorly performing global model. Therefore, we propose a principal gradient-based server aggregation strategy to mitigate conflicting gradients by prioritizing principal directions that benefit all clients while discarding conflict-contributing directions, ultimately reducing the aggregation loss. 
\par In summary, our key contributions are as follows: (i) We propose a novel global loss decomposition in FL, decomposing the global objective into local loss, distribution shift loss, and aggregation loss, which provides an analytical framework to assess the impact of these loss terms on FL performance.  
(ii) 
We provide a novel and practical algorithm, FedLD, which incorporates margin control regularization and a principal gradient-based server aggregation strategy to jointly reduce local loss, distribution shift loss, and aggregation loss.
(iii) We conduct extensive numerical studies on retinal and chest X-ray classification datasets to verify the performance of our algorithm, which outperforms several classic baselines under different levels of data heterogeneity.

\section{Methodology}
\subsection{Problem Setup}
In this work, we address the problem of data heterogeneity in cross-device FL involving a central server and $m$ clients, for a supervised image classification task. Each client $i$ has its own local data distribution, denoted by  $\mathcal{P}_i$. Let $x$ and $y$ indicate the input features and labels extracted from client $i$'s local data distribution $\mathcal{P}_i$,  respectively, such that $(x,y) \sim \mathcal{P}_i(x,y)$. 
Then the objective is to minimize the aggregate loss function $\mathcal{L}(\boldsymbol{w})$, which is formulated as:
\begin{equation}
\label{objective}
\mathcal{L}(\boldsymbol{w})=\sum_{i=1}^m \frac{n_i}{n} \mathcal{L}_i(\boldsymbol{w}),
\end{equation}
where $\mathcal{L}_i(\boldsymbol{w})=\mathbb{E}_{(x, y) \sim \mathcal{P}_i(x, y)}\left[l\left(\boldsymbol{w} ; x,y\right)\right]$ is the empirical loss of client $i$, $n_i$ is the number of samples for client $i$ and $n=\sum_{i=1}^{m} {n_i}$, and $\boldsymbol{w}$ denotes the global model parameter. In data heterogeneity setting, $\mathcal{P}_i \neq \mathcal{P}_j$ for different client $i$ and client $j$. This disparity causes the local model to perform differently across clients, which degrades FL performance or even causes model divergence. 
\subsection{Global Loss Objective Decomposition}
\par To better understand the influence of data heterogeneity on FedAvg in each round, we decompose the loss function in Eq.(\ref{objective}) as follows:
\small{
\begin{equation}  
\label{objective_decomposition}  
\begin{aligned}  
\mathcal{L}(\boldsymbol{w})= & \sum_{i=1}^m \frac{n_i}{n} \mathcal{L}_i(\boldsymbol{w})\\  
=&\underbrace{\sum_{i=1}^{m} \frac{n_i}{n} \mathcal{L}_i\left(\boldsymbol{w}_i\right)}_{\substack{\text{Local loss}\\}}+\underbrace{\sum_{j=1}^m \sum_{i=1}^m \frac{n_j}{n} \frac{n_i}{n}\left(\mathcal{L}_j\left(\boldsymbol{w}_i\right)-\mathcal{L}_i\left(\boldsymbol{w}_i\right)\right)}_{\text{Distribution shift loss}}+\underbrace{\sum_{i=1}^m \frac{n_i}{n}\left(\mathcal{L}(\boldsymbol{w})-\mathcal{L}\left(\boldsymbol{w}_i\right)\right)}_{\text{Aggregation loss}}.  
\end{aligned}  
\end{equation}
}Here $\mathcal{L}_j\left(\boldsymbol{w}_i\right)$ denotes the empirical loss of client $i$'s local model $\boldsymbol{w}_i$ when evaluated on the client $j$'s local dataset. The formula in Eq.(\ref{objective_decomposition}) holds because $\sum_{j=1}^{m}\frac{n_j}{n}\mathcal{L}_j(\cdot) = \mathcal{L}(\cdot)$. We can interpret different terms in Eq.(\ref{objective_decomposition}) as follows: (i) $\mathcal{L}_i\left(\boldsymbol{w}_i\right)$ in the first term denotes the empirical loss of client $i$'s local model $\boldsymbol{w}_i$ trained on its local dataset. We thus refer to the first term as the local loss.  (ii) $\mathcal{L}_j\left(\boldsymbol{w}_i\right)-\mathcal{L}_i\left(\boldsymbol{w}_i\right)$ in the second term denotes the increase in the empirical loss of client $i$'s local model $\boldsymbol{w}_i$ when evaluated on client $j$'s local dataset compared to client $i$'s local dataset. This increase arises from the data distribution shift between client $i$ and client $j$. We thus call the absolute value of the second term the distribution shift loss. (iii) $\mathcal{L}(\boldsymbol{w})-\mathcal{L}\left(\boldsymbol{w}_i\right)$ in the third term denotes the increase in the empirical loss on all samples of local datasets for the global model ($\boldsymbol{w}$, obtained after aggregating local models), compared with local model $\boldsymbol{w}_i$. As this increase comes from the server aggregation operation, we refer to the absolute value of this term as the aggregation loss. 

\begin{figure}[h]  
    \centering  
    \includegraphics[width=0.55\textwidth]{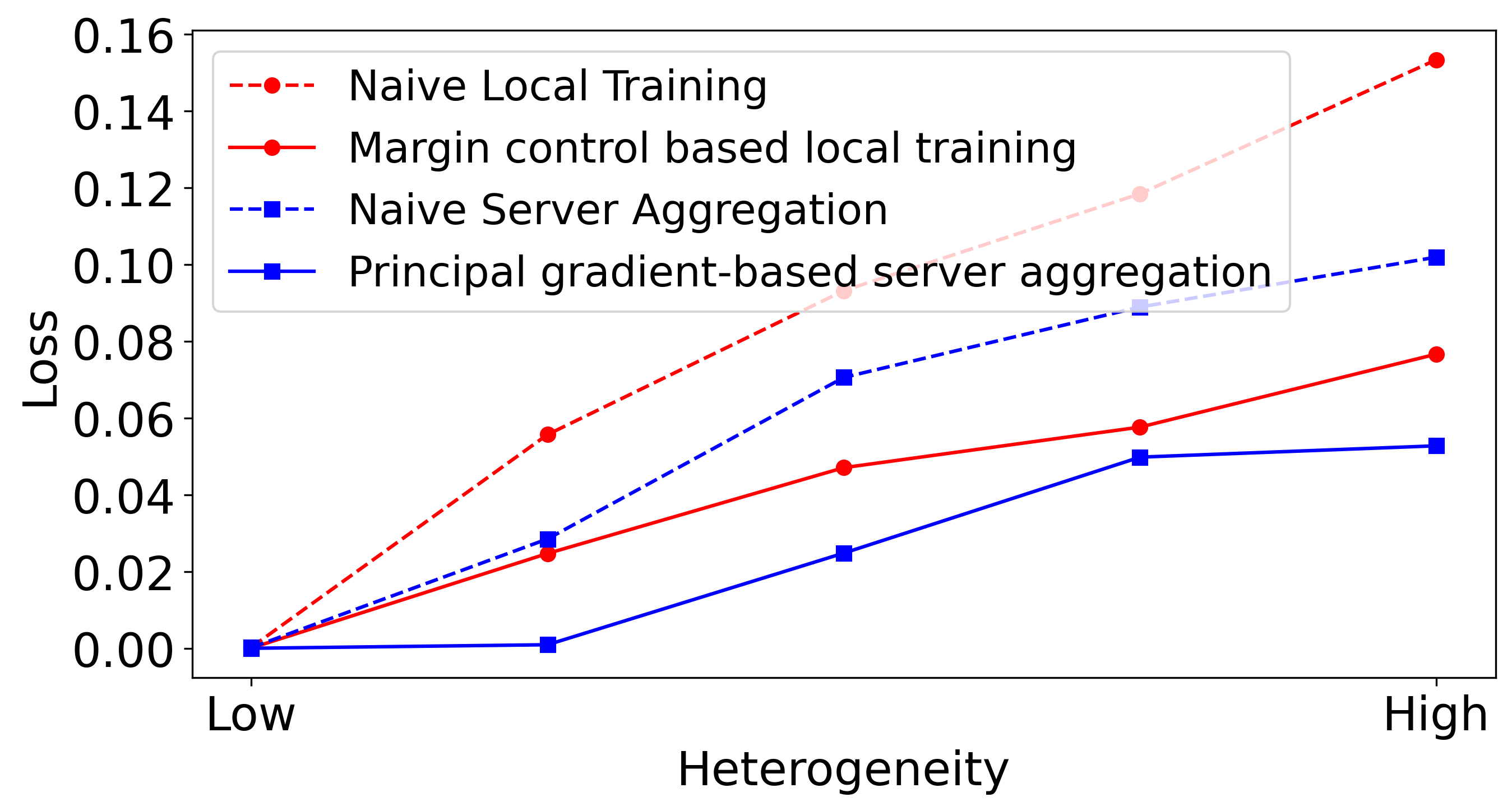}  
    \caption{Distribution shift loss of different local training methods (red lines) and aggregation loss of different server aggregation methods (blue lines) under different levels of heterogeneity in one FL round. 
    }
    \label{fig:comparison}  
\end{figure}

The derived loss decomposition reveals that, in each round, FedAvg only minimizes the local loss (first term in Eq.(\ref{objective_decomposition})) through local training, while ignoring the distribution shift loss and aggregation loss. As shown in Fig. \ref{fig:comparison} (dashed lines), both the distribution shift loss and aggregation loss increase with data heterogeneity. This results in poorer performance and slower convergence of the FedAvg algorithm when facing increased data heterogeneity challenges. This observation motivates us to explore methods that jointly minimize the decomposed three terms in Eq.(\ref{objective_decomposition}). To reduce the distribution shift loss, we need to improve the performance of each local model $\boldsymbol{w}_i$ on the data distributions of other clients during its local training step. To reduce the aggregation loss, we need to develop a more effective way to aggregate local models during server aggregation step. The overview of our method is shown in Fig. \ref{fig:framework}, and we will elaborate these two compenents in the following. Furthermore, the detailed description of our algorithm can be found in supplementary material. 

\subsection{Margin Control in Clients' Local Training}
Shortcut learning refers to a machine learning model's reliance on unstable correlations i.e. shortcut features \cite{geirhos2020shortcut}. It can lead to poor performance when the relationship between the label and the shortcut feature changes \cite{koh2021wilds}. 
This shortcut learning could be a key factor causing a  local model $\boldsymbol{w}_i$ trained on its local dataset to perform worse on other datasets with different distributions. In other words, shortcut learning increases the distribution shift loss in Eq.(\ref{objective_decomposition}). This motivates us to find a method for mitigating shortcut learning in local training to reduce the distribution shift loss. \cite{puli2024don} shows that training with cross entropy loss can lead to the preference for maximizing the margin i.e. range of the logits values, which in turn causes the model to rely more on shortcut features rather than stable features. Motivated by this, we penalize the margin in local training, aiming to reduce the reliance on shortcut features and encourage the model to learn more stable features, thereby reducing the distribution shift loss. Specifically, we use an approach which introduces a margin control regularization term by calculating the $l_2$-norm of the output logits and adding it to the cross entropy loss as follows:
\begin{equation}
\label{marg_log_loss}
\mathcal{L}_{\text {marg-log }}=\mathcal{L}_{C E}\left(y, f_{\boldsymbol{w}}(x)\right)+\lambda \log \left(1+\left\|f_{\boldsymbol{w}}\left(x\right)\right\|_2^2\right),
\end{equation}
where 
$\mathcal{L}_{C E}\left(y, f_{\boldsymbol{w}}(x)\right)=-\sum_{i=1}^C y_i \log \left(\operatorname{softmax}\left(f_{\boldsymbol{w}, i}(x)\right)\right),$ is the standard cross-entropy loss function, and $y = [y_1,\cdots,y_C]^\top, f_{\boldsymbol{w}}(x) = [f_{\boldsymbol{w}, 1}(x),\cdots,f_{\boldsymbol{w}, C}(x)]^\top$.
Here $C$ is the number of classes for our classification task, $y$ is the one-hot vector for the label, and $f_{\boldsymbol{w}}(\cdot)$ is the output logits of the model $\boldsymbol{w}$. Margin control regularization helps to mitigate the reliance on shortcut features and encourages the model to focus on stable features. As shown in Fig. \ref{fig:comparison} (red lines), margin control regularization helps to reduce the distribution shift loss. While $l_2$-norm regularization on output logits was also used in FedSR \cite{nguyen2022fedsr} to align representations from different clients with a common reference, our margin control regularization is motivated by shortcut learning. This allows us to use other regularization techniques, such as evaluating log loss on a margin multiplied by a decreasing function or penalizing large margins by setting thresholds.
\subsection{Principal Gradient-based Server Aggregation}
One of the key factors contributing to the performance degradation of FL in data heterogeneity is the conflicting gradients of local models trained on diverse local datasets  \cite{charles2021large,zhang2023tackling}. Consider a scenario of FL training on two clients with datasets A and B. These datasets are unevenly distributed subsets of the same population, where dataset A predominantly contains "class A" data, and dataset B mainly comprises "class B" data. The models trained on these two datasets may exhibit different or even conflicting judgments, leading to potential conflicts in the gradients of the clients trained on these datasets.  
Naively aggregating the conflicting gradients in FL may lead to a poorly performing global model, thus increasing the aggregation loss in Eq.(\ref{objective_decomposition}). Therefore, we propose a principal gradient-based server aggregation approach to amend these conflicting gradients and force them to follow a direction that maximally benefits all participating clients. The specific steps are:

\noindent\textbf{Step 1: Principal Gradients Construction.} For client $i$, we flatten its local gradients into a vector, denoted as $\boldsymbol{g}_{i} \in \mathbb{R}^{d \times 1}$, where $d$ is the flattened dimension of the local gradients and is usually fairly large for modern deep learning architectures.
All the participating local gradient vectors make up a matrix $\boldsymbol{G}=[\boldsymbol{g}_1,\cdots, \boldsymbol{g}_m]$. Next, we perform singular value decomposition (SVD) on matrix $\boldsymbol{G}$, generating a series of eigenvalues and their corresponding eigenvectors, expressed as: 
\begin{align}
\label{svd_decomposition}
\lambda_{z}, \boldsymbol{v}_{z} & = \text{SVD}_{z}\left(\frac{1}{m} {\boldsymbol{G}} {\boldsymbol{G}^\top}\right),  
\end{align}
where $\lambda_{z}$ and $\boldsymbol{v}_{z}$ represent the $z$-th largest eigenvalue and its corresponding eigenvector, respectively. However, the high dimension of $\boldsymbol{G} \boldsymbol{G}^{\top} \in \mathbb{R}^{d \times d}$ makes naive SVD complex and prohibitive. Thus, we construct a bijection \cite{wang2023pgrad} to reduce computational complexity as follows: 
\begin{align}
\label{svd_decomposition_1}
{\boldsymbol{G}^\top} {\boldsymbol{G}} \boldsymbol{e}_{z} & = \lambda_{z} \boldsymbol{e}_{z} \quad \Longrightarrow \quad {\boldsymbol{G}} {\boldsymbol{G}^\top} {\boldsymbol{G}} \boldsymbol{e}_{z} = \lambda_{z} {\boldsymbol{G}} \boldsymbol{e}_{z} \quad \Longrightarrow \quad \boldsymbol{v}_{z} = {\boldsymbol{G}}
\boldsymbol{e}_{z}.
\end{align}
Here $\boldsymbol{e}_{z}$ represents the $z$-th largest eigenvector of the matrix ${\boldsymbol{G}^\top} {\boldsymbol{G}}$. In this way, we can transform the computation of SVD for $\boldsymbol{G} \boldsymbol{G}^{\top} \in \mathbb{R}^{d \times d}$ in Eq.($\ref{svd_decomposition}$) to the computation of SVD for $\boldsymbol{G}^{\top} \boldsymbol{G} \in \mathbb{R}^{m \times m}$, which is much cheaper, as $m \ll d$. After performing SVD in Eq.(\ref{svd_decomposition_1}), we obtain a set of eigenvectors $\boldsymbol{V}=\{\boldsymbol{v}_{1},\cdots, \boldsymbol{v}_{m}\}$ and consider them as the principal gradients.

\noindent\textbf{Step 2: Local Gradients Revision based on Calibrated Principal Gradients.} These eigenvectors are ordered based on the magnitude of the eigenvalues and are unoriented, since a negative multiple of an eigenvector is also a valid eigenvector. However, we need to determine the directions of these eigenvectors so that they point to the directions that can reduce the loss.  For simplicity, we use the mean of local gradients $\hat{\boldsymbol{g}} =  \frac {1} {m} \sum_i  \boldsymbol{g}_{i}$ as a reference for calibrating the principal gradient direction. Specifically, we adjust the $z$-th largest oriented eigenvector to be positively related to the reference by following the  calibration process: 
\begin{equation}
\bar{\boldsymbol{v}}_z= \begin{cases}\boldsymbol{v}_z, & \text { if }\left\langle \boldsymbol{v}_z, \hat{\boldsymbol{g}}\right\rangle \geq 0 \\ -\boldsymbol{v}_z, & \text { otherwise }\end{cases}.
\end{equation}
\par Next, we select the eigenvectors with the top $L$ largest eigenvalues, i.e. $\{\bar{\boldsymbol{v}}_{1},\cdots, \bar{\boldsymbol{v}}_{L}\}$, to form the principal coordinate system for local gradient projection. For each client $i$, we project its local gradient $\boldsymbol{g}_{i}$ onto this principal coordinate system, with the projection of $\boldsymbol{g}_{i}$ on the $l$-th eigenvector (axis) is calculated as 
$\boldsymbol{g}_{i, l}^{\prime}=\frac{\boldsymbol{g}_i \bar{\boldsymbol{v}}_l}{\left\|\bar{\boldsymbol{v}}_l\right\|\left\|\bar{\boldsymbol{v}}_l\right\|} \bar{\boldsymbol{v}}_l.$
We then aggregate all the projections together into a weighted sum, using the $l$-th eigenvalue $\lambda_l$ as the weight for the $l$-th axis. Additionally, we apply a length correction to the weighted sum by multiplying it by $
\frac{\left\|\boldsymbol{g}_i\right\|}{\| \boldsymbol{g}_{i, l}^{\prime} \|}$:  
\begin{equation}
\label{len_correction}
 \boldsymbol{g}_i^{revise}=\sum_{l=1}^L \frac{\|{\boldsymbol{g}}_i\|} {\|\boldsymbol{g}_{i, l}^{\prime}\|} \frac{\lambda_l}{\|\lambda_l\|}\boldsymbol{g}_{i, l}^{\prime},
\end{equation}
where $\boldsymbol{g}_i^{revise}$ is the revised gradient for the local client $i$. The length correction factor $
\frac{\left\|\boldsymbol{g}_i\right\|}{\| \boldsymbol{g}_{i, l}^{\prime} \|}$ in Eq.(\ref{len_correction}) aims to ensure that the magnitude of the revised local gradient for client $i$ remains the same as the original local gradient $\boldsymbol{g}_i$. This is crucial because a reduced magnitude of the revised local gradient can hinder FL convergence \cite{an2024federated}. 

\noindent\textbf{Step 3: Global Gradient Aggregation.} Last, we aggregate all the revised local gradients $\boldsymbol{g}_i^{revise}$ to construct the global gradients in the server as:
$
\bar{\boldsymbol{g}}=\sum_{i=1}^m \frac{n_i}{n}\boldsymbol{g}_i^{revise}.
$

\begin{remark}
\label{svd_analysis}
To better comprehend the gradient projection based on SVD, let's examine $\frac{1}{m} {\boldsymbol{G}}{\boldsymbol{G}^\top}$ in (\ref{svd_decomposition}) as:  
\begin{equation}
\frac{1}{m} \boldsymbol{G} \boldsymbol{G} ^\top=\frac{1}{m}\left[\boldsymbol{g}_1, \cdots, \boldsymbol{g}_m\right] \otimes\left[\boldsymbol{g}_1, \cdots, \boldsymbol{g}_m\right]=\frac{1}{m} \sum_i \boldsymbol{g}_i \otimes \boldsymbol{g}_i=\frac{1}{m} \sum_i \mathcal{I}_i=-\frac{1}{m} \sum_i \mathcal{H}_i.
\end{equation}
The $\mathcal{I}_i$ and $\mathcal{H}_i$ in the above formulation represent the Fisher Information matrix and Hessian matrix, respectively, and $\otimes$ represents the tensor product in mathematics. The approximation is a positive semi-definite covariance matrix. The eigenvalue $\lambda_{z}$ and eigenvector $v_{z}$ is one-to-one correspondence and arranged based on the size of eigenvalue. For the eigenvalue $\lambda_z$, it is the curvature of the loss in the direction of $\boldsymbol{v}_z$. Since the distribution of the eigenvalues could affect the training behavior, e.g the first-order optimization methods slow down significantly when $\left\{\lambda_z\right\}$ are highly spread out, we can get rid of those insignificant directions and only use the directions with the large curvature. 
Thus, the proposed principal gradient-based server aggregation can mitigate conflicting gradients by prioritizing principal directions that benefit all clients while discarding conflict-contributing directions. The results in Fig. \ref{fig:comparison} (blue lines) also demonstrate the effectiveness of the proposed principal gradient-based server aggregation, as it significantly reduces aggregation loss.

\end{remark}
\section{Experiments}
\subsection{Experimental Setup}

We evaluate our method on two medical image classification datasets: Retina \cite{yan2023label} with 5 clients and COVID-FL \cite{yan2023label}, a real-world federated dataset with 12 clients that exhibits both shifts in label and feature distributions. Following \cite{yan2023label}, we construct different levels of data heterogeneity for Retina by constructing label shifts using Dirichlet distribution with $\alpha$ 100, 1.0, 0.5 for split 1, 2, 3 respectively, i.e., Split-1 (IID), Split-2 (moderate non-IID), and Split-3 (severe non-IID). The ResNet-50 \cite{he2016deep} is adopted in all experiments. We compare our method with FedAvg \cite{mcmahan2017communication}, FedProx \cite{li2020federated},  FedBN \cite{li2021fedbn}, FedPAC \cite{xu2023personalized}, and FedGH \cite{zhang2023tackling}. For optimization, we adopt the SGD optimizer \cite{ruder2016overview} with a learning rate of 0.01 for Retina and 0.005 for COVID-FL. The batch size is set to 50. The number of global communication rounds is set to 200, and the number of local training epochs is set to 1. We choose $\lambda$ values of 0.1, 0.03, and 0.03 for Split-1, Split-2, and Split-3 of Retina, respectively, and 0.01 for COVID-FL. We set $L=0.8m$, where $m$ is the number of selected participating clients in each round. We set the client sampling rate to 1, unless otherwise stated.
\subsection{Evaluation Results}


\begin{table}[h]
  \centering
  \begin{minipage}{0.45\linewidth}
  \caption{The comparison of final test accuracy (\%) of different methods.}  
\label{tab:main_result} 
    \centering
    \begin{tabular}{l*{6}{c}}  
\hline  
Method & \multicolumn{3}{c}{Retina} & COVID-FL \\ \hline  
& Split-1 & Split-2 & \multicolumn{1}{c}{Split-3} & \\ \hline  
FedAvg & 83.63 & 82.26 & 81.13 & 79.86 \\ \hline  
FedProx & 84.17 & 83.53 & 81.20 & 81.88 \\ \hline  
FedBN & 83.91 & 75.91 & 65.25 & 56.34 \\ \hline  
FedPAC & 78.01 & 71.64 & 52.81 & 81.63 \\ \hline  
FedGH & 83.90 & 83.33 & 81.56 & 82.26 \\ \hline  
Ours & \textbf{85.20} & \textbf{83.83} & \textbf{82.30} & \textbf{83.87} \\ \hline  
\end{tabular}  
  \end{minipage}\hfill
  \begin{minipage}{0.45\linewidth}  
    \centering
    \caption{Ablation study. \textbf{Margin} denotes margin control regularization. \textbf{Principal} denotes principal gradient-based server aggregation.}
    \label{tab:ablation_study_component}
    \begin{tabular}{l*{6}{c}}  
\hline  
Margin & Principal & \multicolumn{3}{c}{Retina} \\ \hline  
& & Split-1 & Split-2 & \multicolumn{1}{c}{Split-3}\\ \hline  
$\times$ & $\times$ &83.63 & 82.26 & 81.13 \\ \hline  
$\times$ & $\checkmark$ & 84.76 & 82.67 & 81.66 \\ \hline  
$\checkmark$ & $\times$ & 84.63 & 83.50 & 81.96 \\ \hline  
$\checkmark$ & $\checkmark$ &85.20 & 83.83 & 82.30 \\ \hline  
\end{tabular}  
  \end{minipage}
\end{table}

As demonstrated in Table \ref{tab:main_result}, our method outperforms all compared methods in all datasets with all levels of data heterogeneity, which demonstrates the ability of our method to effectively alleviate the negative impact of data heterogeneity.

\begin{table}[h]  
\centering  
\caption{The comparison of final test accuracy (\%) of different methods on Retina with 50 clients. We apply client sampling with rate 0.1 for FL training.}
\label{tab:number_of_clients_impact}
\begin{tabular}{l*{6}{c}}    
\hline    
Method & FedAvg & FedProx & FedBN & FedPAC & FedGH & Ours \\ \hline
Accuracy (\%) & 67.43 & 69.87 & 68.40 & 66.59 & 70.33 & \textbf{71.30} \\ \hline   
\end{tabular}    
\end{table}

\subsection{Analysis}
\textbf{Ablation Study.}
As demonstrated in Table \ref{tab:ablation_study_component}, both of them can help improve the average test accuracy and the combination of them is able to achieve the most satisfactory model performance, which demonstrates the effectiveness of each component.

\noindent \textbf{Ability of Loss Reduction.}
We simulate different levels of data heterogeneity using the Dirichlet distribution \cite{hsu2019measuring} with the concentration parameter $\alpha \in \{100, 10, 1, 0.1, 0.01\}$ 
on the Retina dataset. First, we train local models with or without margin control regularization. Second, we aggregate the local models trained in the naive way with principal gradient-based aggregation or naive server aggregation. As shown in Fig. \ref{fig:comparison}, margin control regularization significantly reduces distribution shift loss compared with naive local training (red lines), and principal gradient-based aggregation significantly reduces aggregation loss compared with naive server aggregation (blue lines).

\noindent \textbf{Generalization to Multiple Clients.}
We further simulate 50 clients on the Retina dataset using the Dirichlet distribution with the concentration parameter $\alpha = 0.5$. As shown in Table \ref{tab:number_of_clients_impact}, our method also achieves the best result with multiple clients, which demonstrates the generalizability of our method.

\section{Conclusion}
In this paper, we propose a global loss decomposition to understand the impact of data heterogeneity on FL performance, which decomposes the global loss into three terms: local loss, distribution shift loss and aggregation loss. We then propose two strategies, margin control regularization and principal gradient-based server aggregation, to reduce them jointly, thus tackling data heterogeneity in FL. Our loss decomposition provides an analytical tool for analysing the impact of different operations on FL performance, and our proposed margin control regularization and principal gradient-based server aggregation can seamlessly integrate into any FL frameworks. Extensive experiments demonstrate that our algorithm effectively reduces the impact of data heterogeneity on FL performance.


\begin{credits}
\subsubsection{\ackname} This work was supported by National Natural Science Foundation of China (62306253) Early career fund (27204623),  Guangdong Natural Science Fund-General Programme (2024A1515010233), and UCSC  hellman fellowship.

\subsubsection{\discintname} The authors have no competing interests to declare that are relevant to the content of this article.

\end{credits}

%
%
%
\bibliographystyle{splncs04}
\bibliography{Paper-1348}
%




\end{document}


\title{Tackling Data Heterogeneity in Federated Learning via Loss Decomposition (Supplementary Material)}
\author{ }
\institute{ }
\maketitle 
\begin{algorithm}[H]    
  \caption{FedLD}  
  \label{FedLD}
  \SetAlgoLined    
  \SetKwInput{Input}{Input}    
  \SetKwInOut{Output}{Output}    
      
  \Input{number of clients selected in each round $m$, number of communication rounds $T$, number of local epochs $E$, local batch size $B$, learning rate $\eta$, regularization hyperparameter $\lambda$, number of eigenvectors selected $L$, number of local samples for each client $n_i$ and $n=\sum_{i=1}^{m} {n_i}$}    
      
  \Output{global model $w_T$}    
      
  \textbf{Server executes:} 
  Initialize $w_0$\; \\
  \For{each round $t=1,\ldots,T$}{    
    \State {$S_t \leftarrow$ Random selected $m$ local clients}
    \State {Receive local gradients $G = [\boldsymbol{g}_1,\cdots, \boldsymbol{g}_m]$ from $S_t$}
    \State {$\hat{\boldsymbol{g}} =  \frac {1} {m} \sum_{i=1}^{m}  \boldsymbol{g}_{i}$}\\
    \For{$z = 1, \cdots, m$}{
    \State {$\lambda_{z}, \boldsymbol{e}_{z} & = \text{SVD}_{z}\left(\frac{1}{m} {\boldsymbol{G}^\top} {\boldsymbol{G}}\right)$ $\backslash \backslash$ refer to Eq.(4)}
    \State {$\boldsymbol{v}_{z} = \boldsymbol{G} e_z$ $\backslash \backslash$ refer to Eq.(5)}
    \State {$\bar{\boldsymbol{v}}_z= \boldsymbol{v}_z, & \text { if }\left\langle \boldsymbol{v}_z, \hat{\boldsymbol{g}}\right\rangle \geq 0; -\boldsymbol{v}_z, & \text { otherwise }$ $\backslash \backslash$ refer to Eq.(6)}
    }
    \State {Select the principal gradients with the top $L$ largest eigenvalues, $\{\bar{\boldsymbol{v}}_{1},\cdots, \bar{\boldsymbol{v}}_{L}\}$}\\
    \For{$i = 1, \cdots, m$}{
    \For{$l = 1, \cdots, L$}{$\boldsymbol{g}_{i, l}^{\prime}=\frac{\boldsymbol{g}_i \bar{\boldsymbol{v}}_l}{\left\|\bar{\boldsymbol{v}}_l\right\|\left\|\bar{\boldsymbol{v}}_l\right\|} \bar{\boldsymbol{v}}_l.$}
    \State {$ \boldsymbol{g}_i^{revise}=\sum_{l=1}^L \frac{\|{\boldsymbol{g}}_i\|} {\|\boldsymbol{g}_{i, l}^{\prime}\|} \frac{\lambda_l}{\|\lambda_l\|}\boldsymbol{g}_{i, l}^{\prime}$}
    $\backslash \backslash$ refer to Eq.(7)}
    \State {$\bar{\boldsymbol{g}}=\sum_{i=1}^m \frac{n_i}{n}\boldsymbol{g}_i^{revise}$}
    \State {Update the global model $\boldsymbol{w}_{t} \leftarrow \boldsymbol{w}_{t-1}$ with $\bar{\boldsymbol{g}}$}\;  
    }
  \textbf{Clients update($i,\boldsymbol{w}_{t}$):}
     \State {$\boldsymbol{w}_{t}^i \leftarrow \boldsymbol{w}_{t}$}
     \State {$\mathcal{B} \leftarrow\left(\right.$split local dataset of client $i$ into batches of size $\left.B\right)$}\\
     \For{local epoch $e = 1,\cdots,E$}{
     \For{batch $b = \{x,y\}\in \mathcal{B}$}{ \State {$l$ = $l_{C E}\left(y, f_{\boldsymbol{w}_{t}^i}(x)\right)+\lambda \log \left(1+\left\|f_{\boldsymbol{w}_{t}^i}\left(x\right)\right\|_2^2\right)$}
     \State {$w_t^i \leftarrow w_t^i-\eta \nabla \ell$}}
     } 
     \State return $w_t^i$ to server
\end{algorithm} 